\documentclass[french,a4paper]{rnti}
\usepackage{indentfirst}
\usepackage{graphics}
\usepackage{subfigure,graphicx}
\usepackage{fancyhdr}
\usepackage{calc}
\usepackage{multirow}
\usepackage{wrapfig}

\usepackage{lscape}

\usepackage{amssymb,amsmath}
\usepackage{algorithm,algorithmic}
\usepackage{amssymb, amsthm, amsmath, amsfonts}

\newcommand{\conj}{\mathrm{Conj}}

\DeclareMathOperator{\ocap}{\displaystyle{\small{\textcircled{{\scriptsize $\cap$}}}}}
\DeclareMathOperator{\ocup}{\displaystyle{\small{\textcircled{{\scriptsize $\cup$}}}}}

\DeclareMathOperator{\I}{\mathcal{I}}
\DeclareMathOperator{\Pos}{\mathcal{P}}
\DeclareMathOperator{\F}{\mathcal{F}}

\titrecourt{Considérant la dépendance dans la théorie des fonctions de croyance}

\nomcourt{Chebbah et al. }

\titre{Considérant la dépendance dans la théorie des fonctions de croyance}

\auteur{Mouna Chebbah\affil{1}\, \affil{2},
        Mouloud Kharoune\affil{1}\\
        Arnaud Martin\affil{1}, Boutheina Ben Yaghlane\affil{3}}

\affiliation{ 
    \affil{1}UMR 6074 IRISA, Université de Rennes1 / IUT de Lannion, Rue Edouard Branly\\BP 3021, 22302 Lannion cedex\\
Mouloud.Kharoune@univ-rennes1.fr, Arnaud.Martin@univ-rennes1.fr\\
\affil{2}LARODEC, ISG Tunis, 41 Rue de la Liberté, Cité Bouchoucha\\2000 Le Bardo, Tunisie\\
						Mouna.Chebbah@univ-rennes1.fr\\
\affil{3}LARODEC, IHEC Carthage, Carthage Présidence 2016,\\Tunisie\\
boutheina.yaghlane@ihec.rnu.tn}

\resume{La fusion d'informations issues de plusieurs sources cherche à améliorer la prise de décision. La théorie des fonctions de croyance, pour réaliser cette fusion, utilise des règles de combinaison faisant bien souvent l'hypothèse forte de l'indépendance des sources. Cette hypothèse d'indépendance n'est cependant pas formalisée ni vérifiée. 
Nous proposons dans cet article un apprentissage de l'indépendance cognitive de sources d'information permettant de mesurer la dépendance ou l'indépendance. Cette mesure exprimée par une fonction de masse est ensuite intégrée par une approche d'affaiblissement avant de réaliser la combinaison d'informations.}

\summary{In this paper, we propose to learn sources independence in order to choose the appropriate type of combination rules when aggregating their beliefs. Some combination rules are used with the assumption of their sources independence whereas others combine beliefs of dependent sources. Therefore, the choice of the combination rule depends on the independence of sources involved in the combination. 
In this paper, we propose also a measure of independence, positive and negative dependence to integrate in mass functions before the combinaision with the independence assumption.}
\begin{document}
%\textbf{Une réponse aux rapporteurs est jointe à la fin de cet article.}
%%%%%%%%%%%%%%%%%%%%%%%%%%%%%%%%%%%%%%%%%%%%%%%%%%%%%%%%%%%%%%%%%%%%%%%%%%%%%%%%%%%%%%%%%%%%%%%%%%%%%%%%%%%%%%%%%%%%%%%%%%%%%%%
\section{Introduction}% 1. Introduction
%%%%%%%%%%%%%%%%%%%%%%%%%%%%%%%%%%%%%%%%%%%%%%%%%%%%%%%%%%%%%%%%%%%%%%%%%%%%%%%%%%%%%%%%%%%%%%%%%%%%%%%%%%%%%%%%%%%%%%%%%%%%%%%%
\emph{La théorie des fonctions de croyance} issue des travaux de \cite{Dempster67a} et \cite{Shafer76} permet une bonne modélisation des données imprécises et/ou incertaines et offre un outil puissant pour fusionner des informations issues de plusieurs sources. Pour ce faire, les données incertaines et imprécises des différentes sources sont modélisées par des fonctions de masse et combinées afin de mettre en évidence les croyances communes et assurer une prise de décision plus fiable. 

Le choix de la règle de combinaison à appliquer repose sur des hypothèses d'indépendance de sources. En effet, certaines règles de combinaison comme celles de \cite{Dempster67a,Smets90a,Yager87,Dubois88} combinent des fonctions de croyance dont les sources sont supposées indépendantes par contre les règles prudente et hardie proposées par \cite{Denoeux06a} n'exigent pas d'hypothèse d'indépendance. L'indépendance cognitive est une hypothèse fondamentale pour le choix du type de règles de combinaison à appliquer. Les indépendances évidentielle, cognitive et doxastique ont été définies dans la cadre de la théorie des fonctions de croyance.\\

D'une part, les travaux de \cite{BenYaghlane02a, BenYaghlane02b} étudient principalement l'indépendance doxastique des variables.\\
D'autre part, les travaux de \cite{Smets93a} et ceux de \cite{Shafer76} ont défini l'indépendance cognitive des variables par une absence d'implication sur la modification des croyances d'une variable en cas de changement des croyances sur l'autre variable. 
Autrement dit, deux variables sont cognitivement indépendantes si la connaissance de la croyance de l'une n'affecte pas celle de l'autre. Dans la littérature, l'indépendance cognitive des variables est formalisée mais l'indépendance cognitive des sources n'est pas abordée.

Ce papier est focalisé sur l'indépendance cognitive des sources, les indépendances doxastique, cognitive et évidentielle des variables ne sont pas abordées. Nous proposons une approche statistique pour l'estimation de l'indépendance cognitive de deux sources.  
Deux sources sont cognitivement indépendantes si elles ne communiquent pas entre elles et si elles n'ont pas le même corpus de croyance\footnote{Le corpus de croyance est l'ensemble de connaissances ou d'informations acquises par une source.}. La méthode proposée permet d'étudier le comportement général de deux sources et de les comparer pour déceler toute dépendance pouvant exister entre elles. 
Dans le cas de sources dépendantes, nous proposons d'étudier le type de cette dépendance. C'est-à-dire analyser les données de sorte à voir si les sources sont \emph{positivement dépendantes} ou si elles sont \emph{négativement dépendantes}. Cette mesure d'indépendance peut soit guider le choix du type de règles de combinaison, soit être intégrée dans les fonctions de masse afin de justifier l'hypothèse d'indépendance des sources. 

Dans la suite de cet article, nous commençons par rappeler quelques notions de base de la théorie des fonctions de croyance. Ensuite, nous présentons dans la troisième section notre approche statistique d'estimation de l'indépendance ou de la dépendance cognitive. 
Lors de cette méthode, nous proposons d'appliquer l'algorithme de classification non-supervisée sur toutes les informations incertaines de chaque source et de chercher un appariement des clusters. 
L'indépendance des sources est estimée à partir des poids attribués à chaque couple de clusters liés. Dans la quatrième section, si les sources sont dépendantes, une étude de la nature de la dépendance est faite afin de voir si cette dépendance est positive ou négative. 
Cette mesure d'indépendance, dépendance positive et dépendance négative peut être prise en compte dans les fonctions de masse des sources afin de justifier l'hypothèse d'indépendance dans la cinquième section. Finalement, avant de conclure dans la septième section, nous présentons dans la sixième section les expérimentations sur des données générées aléatoirement.  
%%%%%%%%%%%%%%%%%%%%%%%%%%%%%%%%%%%%%%%%%%%%%%%%%%%%%%%%%%%%%%%%%%%%%%%%%%%%%%%%%%%%%%%%%%%%%%%%%%%%%%%%%%%%%%%%%%%%%%%%%%%%%%%%
\section{Théorie des fonctions de croyance} % 2.La théorie des fonctions de croyance
%%%%%%%%%%%%%%%%%%%%%%%%%%%%%%%%%%%%%%%%%%%%%%%%%%%%%%%%%%%%%%%%%%%%%%%%%%%%%%%%%%%%%%%%%%%%%%%%%%%%%%%%%%%%%%%%%%%%%%%%%%%%%%%%
La théorie des fonctions de croyance initialement introduite par \cite{Dempster67a}, formalisée ensuite par \cite{Shafer76} est employée dans des applications de fusion d'informations. Nous présentons ci-dessous quelques principes de base de cette théorie.
%%%%%%%%%%%%%%%%%%%%%%%%%%%%%%%%%%%%%%%%%%%%%%%%%%%%%%%%%%%%%%%%%%%%%%%%%%%%%%%%%%%%%%%%%%%%%%%%%%%%%%%%%%%%%%%%%%%%%%%%%%%%%%%%
\subsection{Principes de base}
%%%%%%%%%%%%%%%%%%%%%%%%%%%%%%%%%%%%%%%%%%%%%%%%%%%%%%%%%%%%%%%%%%%%%%%%%%%%%%%%%%%%%%%%%%%%%%%%%%%%%%%%%%%%%%%%%%%%%%%%%%%%%%%%
Soit un \emph{cadre de discernement} $\Omega=\{\omega_1,\hspace{0.1cm}\omega_2,\hspace{0.1cm}\ldots,\hspace{0.1cm}\omega_n\}$ l'ensemble de toutes les hypothèses exclusives et exhaustives. Le cadre de discernement est aussi \emph{l'univers de discours} d'un problème donné.

L'ensemble $2^\Omega=\{A|A\subseteq\Omega\}=\{\emptyset,\hspace{0.1cm}\omega_1,\hspace{0.1cm}\omega_2,\ldots,\hspace{0.1cm}\omega_n,\hspace{0.1cm}\omega_1\cup\omega_2,\hspace{0.1cm}\ldots,\hspace{0.1cm}\Omega\}$, est l'ensemble de toutes les hypothèses de $\Omega$ ainsi que leurs disjonctions.

Une \emph{fonction de masse} est une fonction de $2^\Omega$ vers l'intervalle $[0,1]$ qui affecte à chaque sous-ensemble une \emph{masse de croyance élémentaire}. Cette fonction de masse fournie par une source d'information\footnote{La source peut être un expert humain, un classificateur, un capteur, \ldots} est une représentation des connaissances incertaines et imprécises. Formellement, une fonction de masse, notée $m^\Omega$, est définie comme suit~:
\begin{equation}%Equation(1)
m^\Omega:2^\Omega \rightarrow\ [0,1]
\end{equation}
tel que:
\begin{eqnarray}%Equation(2)
\sum_{A\subseteq \Omega} m^\Omega(A)=1
\end{eqnarray}

Un sous-ensemble ayant une masse de croyance élémentaire non-nulle est \emph{un élément focal}. La masse d'un élément focal $A$ représente le degré de croyance élémentaire de la source à ce que l'hypothèse vraie soit dans $A$.

%%%%%%%%%%%%%%%%%%%%%%%%%%%%%%%%%%%%%%%%%%%%%%%%%%%%%%%%%%%%%%%%%%%%%%%%%%%%%%%%%%%%%%%%%%%%%%%%%%%%%%%%%%%%%%%%%%%%%%%%%%%%%%%%
\subsection{Combinaison}
%%%%%%%%%%%%%%%%%%%%%%%%%%%%%%%%%%%%%%%%%%%%%%%%%%%%%%%%%%%%%%%%%%%%%%%%%%%%%%%%%%%%%%%%%%%%%%%%%%%%%%%%%%%%%%%%%%%%%%%%%%%%%%%%
Dans le cadre de la théorie des fonctions de croyance, plusieurs règles de combinaison sont proposées pour la fusion d'informations. Les fonctions de masse sont issues de différentes sources et sont définies sur le même ensemble de discernement. La combinaison permet de synthétiser ces différentes informations en vue d'une prise de décision plus fiable. Le choix des règles de combinaison dépend de certaines hypothèses initiales, les opérateurs de type conjonctif peuvent être employés lorsque les sources sont fiables et \emph{indépendantes cognitivement} que nous définirons précisément à la section~\ref{independance}. La combinaison conjonctive s'écrit pour deux fonctions de masse $m^\Omega_1$ et $m^\Omega_2$ et pour tout $A \subseteq \Omega$ par~:
\begin{eqnarray}
\label{conjunctive}
m^\Omega_{\ocap}(A)=m^{\Omega}_1\ocap m^{\Omega}_2(A)= \displaystyle \sum_{B\cap C =A} m^\Omega_1(B)\times m^\Omega_2(C).
\end{eqnarray}
Notons que l'élément neutre pour cette règle est la masse~: $m^\Omega(A)=1$ si $A=\Omega$ et 0 sinon. Lorsque l'hypothèse de fiabilité est trop forte et que l'on ne peut supposer que seule une des sources est fiable, la combinaison disjonctive peut alors être employée toujours sous l'hypothèse d'indépendance cognitive~:
\begin{eqnarray}
\label{disjunctive}
m^\Omega_{\ocup}(A)=m^\Omega_1\ocup m^\Omega_2(A)=\displaystyle \sum_{B\cup C =A} m^\Omega_1(B)\times m^\Omega_2(C).
\end{eqnarray}

Notons que l'élément neutre pour cette règle est la masse~: $m^\Omega(A)=1$ si $A=\emptyset$ et 0 sinon. Bien que les règles disjonctive et conjonctive soient associatives et commutatives, elles ne sont pas idempotentes ce qui justifie leur inefficacité à la fusion d'informations issues de sources dépendantes cognitivement. La plupart des règles de combinaison issues des règles conjonctives et disjonctives, en particulier pour répartir le conflit, supposent que les sources sont indépendantes cognitivement. \cite{Martin10a} en rappelle quelques unes. 

Dans cet article, nous utilisons la moyenne pour la combinaison. Cette règle est choisie parce qu'elle est idempotente et commutative en plus elle combine tous types de fonctions de masse. Toute autre règle de combinaison vérifiant ces critères peut être utilisée.\\
Pour chaque élément focal $A$ des $M$ fonctions de masse, la masse combinée de $A$,\break $m_{Moyenne}^\Omega(A)$, est calculée à partir des $M$ masses de croyance élémentaires $m_{i}^{\Omega}(A)$ comme suit~:
\begin{equation}%Equation(1.31)
\label{Equation(8)}
m^\Omega_{Moyenne}(A)=\frac{1}{M}\sum^{M}_{i=1}m^\Omega_i(A)
\end{equation}
%%%%%%%%%%%%%%%%%%%%%%%%%%%%%%%%%%%%%%%%%%%%%%%%%%%%%%%%%%%%%%%%%%%%%%%%%%%%%%%%%%%%%%%%%%%%%%%%%%%%%%%%%%%%%%%%%%%%%%%%%%%%%%%%
\subsection{Conditionnement et déconditionnement}
%%%%%%%%%%%%%%%%%%%%%%%%%%%%%%%%%%%%%%%%%%%%%%%%%%%%%%%%%%%%%%%%%%%%%%%%%%%%%%%%%%%%%%%%%%%%%%%%%%%%%%%%%%%%%%%%%%%%%%%%%%%%%%%%
Après l'acquisition d'une fonction de masse, une information certaine peut apparaître confirmant que l'hypothèse vraie est (ou n'est pas) dans l'un des sous-ensembles de $2^\Omega$. Dans ce cas, la fonction de masse doit être mise à jour afin de prendre en considération cette nouvelle information certaine. Cette mise à jour est réalisée par l'opérateur de \emph{conditionnement} qui consiste à transférer la masse attribuée à chaque élément focal à son intersection avec l'ensemble certain. Le conditionnement d'une fonction de masse $m^\Omega$ par l'hypothèse $A\subset\Omega$ revient à transférer les masses de croyance de tous les éléments focaux de $m^\Omega$ à leurs intersections avec $A$. La fonction de masse conditionnée $m^\Omega{[A]}$ est donnée par \cite{Smets97c} comme suit:
\begin{equation}
m^\Omega{[A]}(C)=\left\{
\begin{tabular}{ll}
$0$&$\text{for}\hspace{0.1cm}C\not\subseteq A$\\
$\displaystyle{\sum _{B \subseteq A^c}} m^\Omega(C \cup B)$&$\text{for}\hspace{0.1cm}C\subseteq A$\\
\end{tabular}
\right.
\end{equation}
avec $A^c$ le complémentaire de $A$, $A^c=\Omega\setminus\{A\}$.
Notons que le conditionnement sur une hypothèse du même cadre de discernement revient à la combinaison conjonctive de la fonction de masse initiale $m^\Omega$ avec la nouvelle fonction de masse certaine $m^\Omega(A)=1$ sachant que $A$ est une hypothèse certaine, donc $m^{\Omega}_{A}\ocap m^\Omega=m^\Omega{[A]}$.

Le \emph{déconditionnement} est l'opération inverse du conditionnement qui permet de retrouver une fonction de masse la moins informative à partir d'une fonction de masse conditionnée.
Étant donnée $m^{\Omega}{[A]}$, la fonction de masse conditionnellement à $A$, il est difficile de retrouver la fonction de masse originale mais il est possible de retrouver la fonction de masse qui engage le moins (\cite{Hsia91} et \cite{Smets93a}) telle que son conditionnement par $A$ donne $m^{\Omega}{[A]}$. Le déconditionnement de $m^{\Omega}{[A]}$ permet de retrouver $m^\Omega$ comme suit:
\begin{equation}
\label{decon}
\begin{tabular}{ll}
$m^\Omega (C\cup A^c)=m^{\Omega}{[A]}(C)$&$\forall C\subseteq 2^\Omega,\hspace{0.1cm}A^c=\Omega\backslash A$
\end{tabular}
\end{equation}
%%%%%%%%%%%%%%%%%%%%%%%%%%%%%%%%%%%%%%%%%%%%%%%%%%%%%%%%%%%%%%%%%%%%%%%%%%%%%%%%%%%%%%%%%%%%%%%%%%%%%%%%%%%%%%%%%%%%%%%%%%%%%%%%
\subsection{Affaiblissement}
\label{affaiblissement}
%%%%%%%%%%%%%%%%%%%%%%%%%%%%%%%%%%%%%%%%%%%%%%%%%%%%%%%%%%%%%%%%%%%%%%%%%%%%%%%%%%%%%%%%%%%%%%%%%%%%%%%%%%%%%%%%%%%%%%%%%%%%%%%
 \cite{Shafer76} a proposé la procédure d'affaiblissement suivante~:
\begin{eqnarray}
^\alpha m^\Omega(A)&=&\alpha\times m^\Omega(A) \quad \forall A\subset\Omega\\
^\alpha m^\Omega(\Omega)&=&1-\alpha\times(1-m^\Omega(\Omega))
\end{eqnarray}
où $\alpha$ est un facteur d'affaiblissement de $[0,1]$. Cette procédure est généralement employée pour affaiblir les fonctions de masse par la fiabilité $\alpha$ de leurs sources. Cette procédure a pour effet d'augmenter la masse sur l'ignorance $\Omega$. \cite{Smets93a} a justifié cette procédure en considérant que~:
\begin{eqnarray}
m^\Omega{[F]}(A)&=&m^\Omega(A)\\
m^\Omega{[\bar{F}]}(A)&=&m^\Omega(X)
\end{eqnarray}
où $m^\Omega(X)=1$ si $X=\Omega$ et 0 sinon, $F$ et $\bar{F}$ représentent la fiabilité et la non fiabilité et $m^\Omega{[F]}$ est une fonction de masse conditionnellement à la fiabilité $F$. Soit $\F=\{F,\bar{F}\}$ le cadre de discernement correspondant, et $m^{\F}$ la fonction de masse représentant la connaissance sur la fiabilité de la source~:
\begin{equation}
\label{fiab}
\left\{
\begin{tabular}{ll}
 $m^{\F}(F)=\alpha$\\
$m^{\F}(\F)=1-\alpha$. \\ 
\end{tabular}
\right.
\end{equation}
Afin de combiner les deux sources d'informations fournissant les deux fonctions de masse $m^\Omega{[F]}$ et $m^{\F}$, il faut pouvoir les représenter dans le même espace $\Omega \times \F$. Ainsi, nous devons effectuer une {\em extension à vide} sur la fonction de masse $m^{\F}$, opération que l'on note $m^{\F\uparrow\Omega\times\F}$~:
\begin{equation}
m^{\F\uparrow\Omega\times\F} \left(Y\right)=\left\lbrace
\begin{array}{ll}
 m^{\F} \left(X\right)&\text{si } Y=\Omega \times X, \quad X\subseteq\F\\
0 &\text{sinon}
\end{array}
\right.
\end{equation}
Dans le cas de la fonction de masse $m^\Omega{[F]}$, il faut déconditionner~:
\begin{equation}
m^{\Omega}[F]^{\Uparrow\Omega\times\F}((A\times
F)\cup(\Omega\times\overline{F}))=m^\Omega\left[F\right]\left(A\right), \quad A\subseteq\Omega
\end{equation}
Il est ainsi possible d'effectuer la combinaison~:
\begin{equation}
m_{\ocap}^{\Omega\times\F}(B)= m^{\F\uparrow\Omega\times\F} \ocap m^{\Omega}{[F]}^{\Uparrow\Omega\times\F} (B), \quad \forall B\subset \Omega\times\F
\end{equation}
Ensuite il faut marginaliser la fonction de masse obtenue pour revenir dans l'espace $\Omega$~:
\begin{equation}
 m^{\Omega\times\F\downarrow\Omega}
\left(A\right)=\displaystyle{
\sum_{
\left\lbrace
B\subseteq\Omega\times\F | Proj\left(B\downarrow\Omega\right)=A
\right\rbrace}
m_{\ocap}^{\Omega\times\F}(B)
}
\end{equation}
où $Proj\left(Y\downarrow\Omega\right)$ est la projection de $Y$ sur $\Omega$. Nous retrouvons ainsi~:
\begin{equation}
 ^\alpha m^\Omega(A)=m^{\Omega\times\F\downarrow\Omega}\left(A\right)
\end{equation}

\cite{Mercier06a} a proposé une extension de cet affaiblissement en contextualisant le coefficient d'affaiblissement $\alpha$ en fonction des sous-ensembles de $\Omega$. \cite{Smets93a} détaille l'extension à vide, le déconditionnement ainsi que la marginalisation.
%%%%%%%%%%%%%%%%%%%%%%%%%%%%%%%%%%%%%%%%%%%%%%%%%%%%%%%%%%%%%%%%%%%%%%%%%%%%%%%%%%%%%%%%%%%%%%%%%%%%%%%%%%%%%%%%%%%%%%%%%%%%%%%%
\subsection{Transformation pignistique}
%%%%%%%%%%%%%%%%%%%%%%%%%%%%%%%%%%%%%%%%%%%%%%%%%%%%%%%%%%%%%%%%%%%%%%%%%%%%%%%%%%%%%%%%%%%%%%%%%%%%%%%%%%%%%%%%%%%%%%%%%%%%%%%%
La prise de décision dans la théorie des fonctions de croyance est fondée sur des probabilités pignistiques issues de la transformation pignistique proposée par \cite{Smets05b}. Cette transformation calcule une probabilité pignistique à partir des fonctions de masse en vue de prendre une décision. Si un expert fournit une fonction de masse reflétant son avis sur la solution d'un problème précis, la probabilité pignistique reflète la probabilité de chaque hypothèse. La probabilité pignisitique $BetP$ d'un élément $A\in \Omega$ est calculée comme suit:
\begin{eqnarray}
\label{pignistic}
BetP(A)=\sum_{C \in 2^\Omega, C \neq \emptyset} \frac{|A \cap C|}{|C|} \frac{m^\Omega(C)}{1-m^\Omega(\emptyset)}.
\end{eqnarray}
La décision peut être prise suivant le principe du maximum de la probabilité pignistique.
%%%%%%%%%%%%%%%%%%%%%%%%%%%%%%%%%%%%%%%%%%%%%%%%%%%%%%%%%%%%%%%%%%%%%%%%%%%%%%%%%%%%%%%%%%%%%%%%%%%%%%%%%%%%%%%%%%%%%%%%%%%%%%%%
\subsection{Classification non-supervisée}
%%%%%%%%%%%%%%%%%%%%%%%%%%%%%%%%%%%%%%%%%%%%%%%%%%%%%%%%%%%%%%%%%%%%%%%%%%%%%%%%%%%%%%%%%%%%%%%%%%%%%%%%%%%%%%%%%%%%%%%%%%%%%%%
Nous proposons ici d'utiliser un algorithme de classification non-supervisée de type $C$-moyenne, utilisant une distance sur les fonctions de masse définie par  \cite{Jousselme01a} comme proposé par \cite{BenHariz06}, \cite{Chebbah12a} et \cite{Chebbah12b}. 
Soit un ensemble $T$ contenant $n$ objets $o_i:1\leq i\leq n$ à classifier dans $C$ clusters. Les valeurs des $o_i$ sont des fonctions de masse $m^\Omega_i$ définies sur un cadre de discernement $\Omega$. Le but est de classifier les $n$ fonctions de masse valeurs des objets $o_i$. 
Les fonctions de masse $m^\Omega_i$ sont fournies par la même source, c'est-à-dire qu'une même source a attribué les valeurs des objets $o_i$. Appliquer un algorithme de classification non-supervisée sur ces fonctions de masse revient à regrouper les fonctions de masse ayant des éléments focaux non contradictoires.
Une mesure de dissimilarité $D(o_i,Cl_k)$ permet de mesurer la dissimilarité entre un objet $o_i$ et un cluster $Cl_k$ comme suit~:
\begin{equation}
D(o_i,Cl_k)=\frac{1}{n_k}\sum_{j=1}^{n_k}d(m_{i}^{\Omega},m_j^{\Omega})
\end{equation}
%and 
\begin{eqnarray}
\label{JDist}
d(m_1^{\Omega},m_2^{\Omega})=\sqrt{\frac{1}{2}(m_1^{\Omega}-m_2^{\Omega})^t\underline{\underline{D}}(m_1^{\Omega}-m_2^{\Omega})}, \\
\underline{\underline{D}}(A,B)=\left\{
\begin{tabular}{ll}
$1$& si $A=B=\emptyset$\\
$\frac{| A\cap B|}{| A\cup B|}$&$ \forall A,B \in 2^{\Omega}$\\
\end{tabular}
\right.
\end{eqnarray}

La dissimilarité d'un objet $o_i$ et un cluster $Cl_k$ est définie par la moyenne des distances entre la fonction de masse $m^\Omega_i$ valeur de cet objet et toutes les $n_k$ fonctions de masse valeurs des $o_j:1\leq j\leq n_k$ objets contenus dans le cluster $Cl_k$.
Chaque objet est affecté au cluster qui lui est le plus similaire (ayant une valeur de dissimilarité minimale) de manière itérative jusqu'à ce qu'une répartition stable soit obtenue.\\
 \indent \`A la fin de la classification non-supervisée, $C$ clusters contenant chacun un certain nombre d'objets sont obtenus.
Dans cet article nous supposons que le nombre de clusters $C$ soit égal à la cardinalité du cadre de discernement ($C=|\Omega|$) puisque $\Omega$ représente les classes possibles dans un problème de classification. 
%Il existe des approches permettant de définir automatiquement le nombre de clusters qui peuvent être appliquées.
L'utilisation de cette mesure de dissimilarité pour la classification assure le regroupement des objets dont les valeurs (fonctions de masse) ne sont pas contradictoires, c'est-à-dire les éléments focaux sont compatibles et intersectent.\\
\indent Notons que l'algorithme de classification non-supervisée est fondée sur une distance sur les fonctions de masse et non pas sur des centres mobiles comme proposé par \cite{BenHariz06} parce que la distance est fortement sensible au nombre d'éléments focaux des centres. Il suffit d'avoir des  fonctions de masse avec différents éléments focaux dans le même cluster pour que la fonction de masse combinée (le centre) ait beaucoup d'éléments focaux.
%%%%%%%%%%%%%%%%%%%%%%%%%%%%%%%%%%%%%%%%%%%%%%%%%%%%%%%%%%%%%%%%%%%%%%%%%%%%%%%%%%%%%%%%%%%%%%%%%%%%%%%%%%%%%%%%%%%%%%%%%%%%%%%%
\section{Indépendance}
\label{independance}
%%%%%%%%%%%%%%%%%%%%%%%%%%%%%%%%%%%%%%%%%%%%%%%%%%%%%%%%%%%%%%%%%%%%%%%%%%%%%%%%%%%%%%%%%%%%%%%%%%%%%%%%%%%%%%%%%%%%%%%%%%%%%%%
L'indépendance a été introduite en premier dans le cadre de la théorie des probabilités pour modéliser l'indépendance statistique des évènements. Avec les probabilités, deux évènements $A$ et $B$ sont indépendants si $P(A\cap B)=P(A)\times P(B)$ ou encore si $P(A | B)=P(A)$.

Les fonctions de masse peuvent être perçues comme des probabilités subjectives fournies par des sources s'exprimant sur un problème étant donné un ensemble de connaissances ou d'informations appelé \emph{corpus de croyance}. Dans le cas d'une hypothèse \emph{d'indépendance cognitive} des sources, les corpus de croyance doivent être distincts et aucune communication entre les sources n'est tolérée.

 \cite{Shafer76} définit l'indépendance cognitive des variables comme étant le non  changement des croyances de l'une des variables si une nouvelle croyance élémentaire apparaît sur l'autre. Il définit également l'indépendance évidentielle, deux variables sont évidentiellement indépendantes par rapport à une fonction de masse si cette fonction de masse peut être retrouvée en combinant les fonctions de masse de ces variables.\\
 
\textbf{Définition 1.} "Two frames of discernment may be called cognitively independent with respect to the evidence if new evidence that bears on only one of them will not change the degrees of support for propositions discerned by the other"\footnote{Deux cadres de discernement sont cognitivement indépendants par rapport à une croyance si toute nouvelle croyance apparaissant sur l'un n'affecte pas la croyance sur l'autre.} (\cite{Shafer76}, page 149).\\

\textbf{Définition 2.} "Two frames of discernment are evidentially independent with respect to a support function if that support function could be obtained by combining evidence that bears on only one of them with evidence that bears on only the other"\footnote{Deux cadres de discernement sont évidentiellement indépendants par rapport à une fonction de masse si cette fonction de masse est obtenue en combinant les croyances sur ces cadres de discernement.} (\cite{Shafer76}, page 149).\\

Dans ce papier, nous nous intéressons à l'indépendance des sources et non pas celle des variables.\\

\textbf{Définition 3.} Deux sources sont cognitivement indépendantes si elles ne communiquent pas et si leurs corpus de croyance sont distincts. \\

Nous définissons la dépendance cognitive comme étant la ressemblance du comportement général de deux sources dû à une dépendance de connaissances ou à une éventuelle communication entre elles. Nous proposons une démarche statistique afin d'étudier l'indépendance cognitive de deux sources. Le but étant soit de tenir compte de cette indépendance dans les fonctions de masse comme détaillé dans la section \ref{IntégrationInd}, soit faire les hypothèses adaptées pour le choix du type de règles de combinaison à appliquer. Nous introduisons d'abord la mesure d'indépendance de deux sources $S_1$ et $S_2$, notée $I_d(S_1,S_2)$, comme étant l'indépendance de $S_1$ de $S_2$. Cette mesure vérifie les axiomes suivants~:
\begin{enumerate}
\item{Non-négative~:} L'indépendance d'une source $S_1$ de $S_2$, $I_d(S_1,S_2)$ est une valeur qui est, soit nulle si $S_1$ est complètement dépendante de $S_2$, soit strictement positive. 
\item{Normalisée~:} $I_d(S_1,S_2)\in[0,1]$, si $I_d$ est nulle alors $S_1$ est complètement dépendante de $S_2$. Si $I_d=1$, alors $S_1$ est complètement indépendante de $S_2$ autrement c'est un degré de $]0,1[$.
\item{Non-symétrique~:} Si $S_1$ est indépendante de $S_2$, cela n'implique pas forcement que $S_2$ soit indépendante de $S_1$. $S_1$ et $S_2$ peuvent être simultanément indépendantes avec des degrés d'indépendance égaux ou différents.
\item{Identité~:} $I_d(S_1,S_1)=0$, toute source est complètement dépendante d'elle même.
\end{enumerate}
Si deux sources $S_1$ et $S_2$ sont dépendantes, alors des éléments focaux similaires sont choisis pour s'exprimer sur des objets similaires. L'approche proposée dans ce papier est une approche statistique pour mesurer le degré d'indépendance cognitive. Nous proposons ainsi de classifier toutes les fonctions de masse des deux sources et de comparer les clusters obtenus. La classification non supervisée regroupe les objets ayant pour valeurs des fonctions de masse similaires. 

Si les clusters des deux sources sont similaires, c'est que les sources ont tendance à choisir des éléments focaux similaires voire non contradictoires pour les mêmes objets, alors il est fort probable qu'elles soient dépendantes. Si les clusters sont fortement liés c'est-à-dire qu'ils contiennent des fonctions de masse relatives aux mêmes objets, alors les sources sont dépendantes cognitivement. 
%%%%%%%%%%%%%%%%%%%%%%%%%%%%%%%%%%%%%%%%%%%%%%%%%%%%%%%%%%%%%%%%%%%%%%%%%%%%%%%%%%%%%%%%%%%%%%%%%%%%%%%%%%%%%%%%%%%%%%%%%%%%%%%%
\subsection{Appariement des clusters}
\label{appariement}
%%%%%%%%%%%%%%%%%%%%%%%%%%%%%%%%%%%%%%%%%%%%%%%%%%%%%%%%%%%%%%%%%%%%%%%%%%%%%%%%%%%%%%%%%%%%%%%%%%%%%%%%%%%%%%%%%%%%%%%%%%%%%%%
 
Dans de nombreuses applications, plusieurs sources s'expriment sur la même problématique et fournissent différentes fonctions de masse comme valeurs aux mêmes objets. L'algorithme de classification non-supervisée est appliqué aux fonctions de masse de chaque source séparément et puis ces clusters doivent être comparés dans le but de voir s'il y a un lien entre eux. Plus les liens entre ces clusters sont forts plus les sources ont tendance à être dépendantes.
\indent Soient deux sources $S_1$ et $S_2$, fournissant chacune $n$ fonctions de masse pour les mêmes objets. Ceci exprime le fait que la fonction de masse $m^\Omega_i$ fournie par $S_1$ et celle fournie par $S_2$ se réfèrent au même objet $o_i$. 
Après avoir classifié les fonctions de masse de $S_1$ et $S_2$, la matrice de correspondance des clusters $M$ est obtenue par~:
\begin{eqnarray}
M_1= \begin{pmatrix}
\beta^1_{1,1}&\beta^1_{1,2}&\ldots& \beta^1_{1,C}\\ 
\ldots&\ldots&\ldots&\ldots\\
\beta^1_{k,1}&\beta^1_{k,2}&\ldots& \beta^1_{k,C}\\   
\ldots&\ldots&\ldots&\ldots\\
\beta^1_{C,1}&\beta^1_{C,2}&\ldots& \beta^1_{C,C}\\  
\end{pmatrix}
& \quad \mbox{and} \quad
M_2= \begin{pmatrix}
\beta^2_{1,1}&\beta^2_{1,2}&\ldots& \beta^2_{1,C}\\ 
\ldots&\ldots&\ldots&\ldots\\
\beta^2_{k,1}&\beta^2_{k,2}&\ldots& \beta^2_{k,C}\\   
\ldots&\ldots&\ldots&\ldots\\
\beta^2_{C,1}&\beta^2_{C,2}&\ldots& \beta^2_{C,C}\\  
\end{pmatrix}
\end{eqnarray}
avec
\begin{equation}
\beta^i_{k_i,k_j}=\frac{|Cl^i_{k_i}\cap Cl^j_{k_j}|}{|Cl^i_{k_i}|}
\end{equation}
Notons que $\beta^i_{k_i,k_j}$ est la similarité des clusters $Cl^i_{k_i}$ de $S_i$ et $Cl^j_{k_j}$ de $S_j$ par rapport à $S_i$ avec $\{i,j\}\in \{1,2\}$ et $i\neq j$.

La similarité de deux clusters est la proportion des objets classés simultanément dans $Cl^i_{k_i}$ et $Cl_{k_j}^j$ par rapport à la cardinalité du cluster de la source référente. Si par exemple, le cluster $Cl^1_1$ de $S_1$ contient les $5$ objets $\{o_1,o_5,o_9,o_{12},o_{15}\}$ et le cluster $Cl^2_5$ de $S_2$ contient les $3$ objets $\{o_1,o_2,o_9\}$. 
$\beta^1_{1,5}$ est la similarité de $Cl^2_5$ de $S_2$ par rapport à $Cl^1_1$ de $S_1$ représentant la proportion des objets de $Cl^2_5$ qui sont aussi classés dans $Cl^1_1$ de $S_1$ donc $\beta^1_{1,5}=\frac{2}{5}$ et $\beta^2_{1,5}=\frac{2}{3}$. Notons que l'algorithme de classification non-supervisée est appliqué sur 
les mêmes objets ayant différentes valeurs puisqu'elles sont fournies par deux sources différentes $S_1$ et $S_2$. 
Les matrices $M_1$ et $M_2$ sont ainsi différentes puisque les clusters de $S_1$ et ceux de $S_2$ sont différents. Notons également que d'autres coefficients de similarité peuvent aussi être utilisés.\\
Une fois les deux matrices de correspondances $M_1$ et $M_2$ calculées, une correspondance entre les clusters est établie suivant l'algorithme \ref{cluster matching}. Chaque cluster est lié au cluster qui lui est le plus similaire, ayant le $\beta$ maximal, en vérifiant que deux clusters de la même source ne peuvent pas être liés au même cluster de l'autre source. Par exemple, nous ne pouvons pas avoir les deux clusters $Cl^1_1$ et $Cl^1_3$ de $S_1$ liés au même cluster $Cl^2_3$ de $S_2$. Pour déterminer la correspondance des clusters pour chaque source, les deux clusters ayant la plus grande similarité $\beta$ sont reliés et écartés de l'ensemble des clusters à apparier.
La recherche de correspondances des clusters est faite pour les deux sources. Deux correspondances différentes peuvent être obtenues pour les deux sources.
\begin{algorithm}[h]
\caption{Appariement de clusters}
\label{cluster matching}
\begin{algorithmic}[1]
\REQUIRE $M$ matrice de correspondances.
\WHILE {$M$ est non vide}
   \STATE Rechercher le maximum de $M$ ainsi que les indices $l$ et $c$ du maximum.
	 \STATE Apparier les clusters $l$ et $c$.
	 \STATE Enlever la ligne $l$ et la colonne $c$ de $M$.
\ENDWHILE
\RETURN Un appariement de clusters.
\end{algorithmic}
\end{algorithm}
%%%%%%%%%%%%%%%%%%%%%%%%%%%%%%%%%%%%%%%%%%%%%%%%%%%%%%%%%%%%%%%%%%%%%%%%%%%%%%%%%%%%%%%%%%%%%%%%%%%%%%%%%%%%%%%%%%%%%%%%%%%%%%%
\subsection{Indépendance des clusters}
%%%%%%%%%%%%%%%%%%%%%%%%%%%%%%%%%%%%%%%%%%%%%%%%%%%%%%%%%%%%%%%%%%%%%%%%%%%%%%%%%%%%%%%%%%%%%%%%%%%%%%%%%%%%%%%%%%%%%%%%%%%%%%%
Une fois la correspondance des clusters établie, une fonction de masse définissant l'indépendance de chaque couple de cluster est déduite. Ceci revient à avoir un agent ayant les correspondances des clusters ($k_{i},k_{j}$) avec les similarités correspondantes $\beta^{i}_{k_i,k_j}$ comme corpus de croyance pour s'exprimer sur l'indépendance de ces clusters. Pour résumer, nous supposons que les deux sources $S_1$ et $S_2$ fournissent $n$ fonctions de masse comme valeurs aux $n$ objets $o_i$. Les fonctions de masse des deux sources sont classifiées chacune à part en utilisant l'algorithme de classification détaillé dans la section~\ref{appariement}. Différents $C$ clusters sont le résultat de classification des fonctions de masse de $S_1$ et ceux de $S_2$.
Après appariement de clusters, les clusters de $S_1$ sont liés aux clusters de $S_2$ qui leur sont similaires et ceux de $S_2$ sont également liés aux clusters de $S_1$ les plus similaires. Différents appariements sont obtenus pour $S_1$ et $S_2$.
Rappelons que le  but est d'estimer l'indépendance cognitive des sources à partir d'un ensemble de fonctions de masse fourni par chacune.

Dans cette section, nous définissons l'indépendance de chaque couple de clusters liés ($k_1,k_2$) comme une fonction de masse définie sur le cadre de discernement $\I=\{\bar{I},I\}$, où $\bar{I}$ représente la dépendance et $I$ l'indépendance:
\begin{equation}
\label{eq35}
\left\{
\begin{tabular}{ll}
%$\displaystyle{m^{\Omega_I}_{{k_1},{k_2}}}(\emptyset)=0$\\
$m^{\I}_{{k_i},{k_j}}(I)=\alpha^i_{{k_i},{k_j}} \, (1-\beta^i_{{k_i},{k_j}})$\\
$m^{\I}_{{k_i},{k_j}}(\bar{I})=\alpha^i_{{k_i},{k_j}} \, \beta^i_{{k_i},{k_j}}$\\
$m^{\I}_{{k_i},{k_j}}(\I)=1-\alpha^i_{{k_i},{k_j}}$\\
\end{tabular}
\right.
\end{equation}
Le coefficient $\alpha^{i}_{{k_i},{k_j}}$ est un facteur d'affaiblissement utilisé pour tenir compte du nombre d'objets contenus dans les clusters de la source référente. Si deux clusters contenant très peu de fonctions de masse sont reliés et que deux autres clusters contenant beaucoup plus d'objets sont aussi reliés avec un même degré de similarité, les fonctions de masse des deux couples de clusters ne doivent pas avoir un même poids. Bien que ce facteur ne soit pas encore défini, il dépend du nombre d'objets dans le cluster de la source référente ainsi que le nombre total d'objets (fonctions de masse).\\
\indent Une fonction de masse est définie pour chaque couple de clusters appariés pour chacune des sources. Pour avoir une fonction de masse sur l'indépendance globale de chaque source, toutes ces fonctions de masse sont combinées avec la moyenne. Pour résumer, les $C$ clusters de $S_1$ sont appariés aux $C$ clusters de $S_2$, une fonction de masse est ainsi obtenue pour chaque couple de clusters reliés afin de refléter leur degré d'indépendance. $C$ fonctions de masse sont alors obtenues pour chaque source. Notons que toute autre règle\footnote{La règle doit permettre de combiner tous types de fonctions de masse (la règle prudente de \cite{Denoeux08a} ne peut pas être utilisée dans ce cas puisqu'elle est limitée à la combinaison des fonctions de masse non dogmatiques).} idempotente et commutative peut être utilisée pour combiner ces fonctions de masse. L'idempotence de la règle est exigée parce que l'indépendance des clusters appariés peut être la même, l'indépendance de la source doit aussi être égale à l'indépendance des clusters dans ce cas. La règle de combinaison doit également être commutative puisqu'elle mixe $C$ fonctions de masse à la fois.  La combinaison de ces $C$ fonctions de masse est une fonction de masse $m^{\I}_i$ décrivant l'indépendance globale de la source $S_i$ par rapport à $S_j$:
\begin{equation}
\label{ind}
\left\{
\begin{tabular}{l}
%$\displaystyle{m^{\Omega_I}_{{k_1},{k_2}}}(\emptyset)=0$\\
$m^{\I}_i(I)=\frac{1}{C}\displaystyle{\sum_{k_i=1}^{C}}m^{\I}_{k_i,k_j}(I)$\\
$m^{\I}_i(\bar{I})=\frac{1}{C}\displaystyle{\sum_{k_i=1}^{C}}m^{\I}_{k_i,k_j}(\bar{I})$\\
$m^{\I}_i(\bar{I}\cup I)=\frac{1}{C}\displaystyle{\sum_{k_i=1}^{C}}m^{\I}_{k_i,k_j}(\bar{I}\cup I)$\\
\end{tabular}
\right.
\end{equation}
ou encore:
\begin{equation}
\label{ind2}
\left\{
\begin{tabular}{l}
%$\displaystyle{m^{\Omega_I}_{{k_1}{k_2}}}(\emptyset)=0$\\
$m^{\I}_i(I)=\frac{1}{C}\displaystyle{\sum_{k_i=1}^{C}}\alpha^i_{k_i,k_j}(1-\beta^i_{k_i,k_j})$\\
$m^{\I}_i(\bar{I})=\frac{1}{C}\displaystyle{\sum_{k_i=1}^{C}}\alpha^i_{k_i,k_j}\beta^i_{k_i,k_j}$\\
$m^{\I}_i(\bar{I}\cup I)=\frac{1}{C}\displaystyle{\sum_{k_i=1}^{C}}(1-\alpha^i_{k_i,k_j})$\\
\end{tabular}
\right.
\end{equation}
 Les probabilités pignistiques calculées à partir de la fonction de masse combinée permettent la prise de décision sur l'indépendance des sources. L'indépendance de la source $S_1$ de la source $S_2$, $I_d(S_1,S_2)$ n'est autre que la probabilité pignistique de $I$, $I_d(S_1,S_2)=BetP(I)$ et $\bar{I_d}(S_1,S_2)=BetP(\bar{I})$ ce qui revient à écrire $I_d$ comme suit~:

\begin{equation}
\left\{
\begin{tabular}{l}
$I_d(S_i,S_j)=\frac{1}{C}\displaystyle{\sum_{k_i=1}^{C}}[\alpha^i_{{k_i},{k_j}} \, \beta^i_{{k_i},{k_j}}+\frac{1}{2}(1-\alpha^i_{{k_i},{k_j}})]$\\
$\bar{I_d}(S_i,S_j)=\frac{1}{C}\displaystyle{\sum_{k_i=1}^{C}}[\alpha^i_{{k_i},{k_j}} \, (1-\beta^i_{{k_i},{k_j}})+\frac{1}{2}(1-\alpha^i_{{k_i},{k_j}})]$
\end{tabular}
\right.
\end{equation}
Si $I_d(S_1,S_2)<\bar{I_d}(S_1,S_2)$, alors $S_1$ est dépendante de $S_2$, dans le cas contraire $S_1$ est indépendante de $S_2$. $I_d$ n'est pas forcement symétrique, c'est-à-dire que le cas où $I_d(S_1,S_2)\neq I_d(S_2,S_1)$ peut être fréquent puisque la correspondance des clusters est différente pour $S_1$ et $S_2$, ainsi ces fonctions de masse d'indépendance des clusters liés sont aussi différentes. Cette propriété permet par exemple de mettre en évidence une indépendance de $S_1$ par rapport à $S_2$ et une indépendance de $S_2$ par rapport à $S_1$.
%%%%%%%%%%%%%%%%%%%%%%%%%%%%%%%%%%%%%%%%%%%%%%%%%%%%%%%%%%%%%%%%%%%%%%%%%%%%%%%%%%%%%%%%%%%%%%%%%%%%%%%%%%%%%%%%%%%%%%%%%%%%%%%
\section{Dépendance positive ou négative}
%%%%%%%%%%%%%%%%%%%%%%%%%%%%%%%%%%%%%%%%%%%%%%%%%%%%%%%%%%%%%%%%%%%%%%%%%%%%%%%%%%%%%%%%%%%%%%%%%%%%%%%%%%%%%%%%%%%%%%%%%%%%%%%
La mesure $I_d(S_1,S_2)$ informe sur l'indépendance ou {\em a contrario} la dépendance de la source $S_1$ par rapport à la source $S_2$ permettant par exemple de choisir la règle de combinaison à utiliser ou encore intégrer cette information dans ses fonctions de masse. Quant au moins l'une des sources $S_1$ ou $S_2$ est dépendante de l'autre ($I_d(S_1,S_2)<\bar{I_d}(S_1,S_2)$ ou $I_d(S_2,S_1)<\bar{I_d}(S_2,S_1)$), il est alors préférable d'utiliser les règles de \cite{Denoeux06a,Boubaker13,Elouedi98} sinon les règles de \cite{Dubois88, Martin07a, Murphy00a, Smets94a, Yager87} permettent par exemple de redistribuer la masse de l'ensemble vide. 
Dans le cas de sources dépendantes $I_d$ n'est pas suffisante pour indiquer le type de la dépendance. 

Deux sources dépendantes peuvent être positivement ou négativement dépendantes. Si $S_1$ est dépendante de $S_2$, elle peut soit avoir les mêmes croyances si elle lui est positivement dépendante soit avoir des croyances contradictoires si elle lui est négativement dépendante.

Si par exemple $I_d(S_1,S_2)<\bar{I_d}(S_1,S_2)$, alors $S_1$ est dépendante de $S_2$ ce qui signifie que les clusters de $S_1$ ressemblent aux clusters de $S_2$. Nous définissons une mesure de conflit entre les clusters de $S_1$ et $S_2$ quantifiant cette dépendance que nous qualifions de positive ou négative. Si les clusters liés ne sont pas conflictuels alors $S_1$ est positivement dépendante de $S_2$ sinon elle est négativement dépendante. Nous définissons alors le conflit entre les deux clusters dépendants $Cl^i_{k_i}$ et $Cl^j_{k_j}$ ($\{i,j\}\in \{1,2\}$ et $i\neq j$) à partir de la moyenne des distances entre les fonctions de masse des objets en commun~:

\begin{equation}
\left\{
\begin{tabular}{ll}
\label{eqconf}
$Conf(Cl^i_{k_i},Cl^j_{k_j})=\frac{1}{|Cl^i_{k_i}\cap Cl^j_{k_j}|}\displaystyle{\sum_{l\in E(Cl^i_{k_i},Cl^j_{k_j})}}d(m_{l}^{\Omega,i},m_l^{\Omega,j})$&$\text{si }|Cl^i_{k_i}\cap Cl^j_{k_j}|\neq 0$\\
$1$&$\text{sinon}$\\
\end{tabular}
\right.
\end{equation}
avec 
\begin{equation}
E(Cl^i_{k_i},Cl^j_{k_j})=\{l\in[1,n], n=|Cl^i_{k_i}\cap Cl^j_{k_j}|,m_{l}^{\Omega,i}\in Cl^i_{k_i} et\hspace{0.1cm}m_{l}^{\Omega,j}\in Cl^j_{k_j}\}
\end{equation}

Cette mesure de conflit est la moyenne des conflits entre les objets contenus dans les clusters $Cl^i_{k_i}$ et $Cl^j_{k_j}$ ce qui explique le fait de ne considérer que les objets communs. Il ne peut s'agir de conflit entre sources que lorsqu'elles s'expriment sur les mêmes problèmes c'est à dire les mêmes objets.
Le conflit est calculé pour chaque couple de clusters liés. Alors si $Cl^1_1$ de $S_1$ est lié à $Cl^2_5$ de $S_2$ et que $S_1$ est dépendante de $S_2$ alors le conflit entre $Cl^1_1$ et $Cl^2_5$ est la moyenne des distances des fonctions de masse relatives aux objets qui sont à la fois dans $Cl^1_1$ et $Cl^2_5$.
Une fonction de masse définie sur le cadre de discernement $\Pos=\{I, P,\bar{P}\}$ (où $P$ représente la dépendance positive et $\bar{P}$ la dépendance négative) décrivant cette dépendance est obtenue pour chaque couple de clusters~:
\begin{equation}
\label{eqdep}
\left\{
\begin{tabular}{ll}
%$m^{\Pos}_{{k_1},{k_2}}(\emptyset)=0$\\
$m^{\Pos}_{{k_i},{k_j}}[\bar{I}](P )=1-Conf(Cl^i_{k_i},Cl^j_{k_j})$\\
$m^{\Pos}_{{k_i},{k_j}}[\bar{I}](\bar{P} )=Conf(Cl^i_{k_i},Cl^j_{k_j})$\\
%$m^{\Pos}_{{k_i},{k_j}}[\bar{I}](P\cup \bar{P} )=0$\\
\end{tabular}
\right.
\end{equation}
Notons que le conflit entre les clusters reflète la contradiction entre ces clusters. Puisque les clusters de chaque source groupent des fonctions de masse ayant les éléments focaux non contradictoires, alors le conflit mesuré par l'équation \eqref{eqconf} compare les clusters en mesurant la contradiction entre des éléments focaux des fonctions de masse des deux clusters. Plus le conflit est important, plus les sources sont dépendantes négativement mais par contre moins il est important plus les sources sont dépendantes positivement. 

Notons que ces fonctions de masse sont conditionnelles puisque la dépendance positive ou négative des clusters n'est mesurée qu'avec une forte hypothèse de dépendance des clusters liés. L'hypothèse de dépendance ou encore de non indépendance des clusters explique le fait que les fonctions de masse de l'équation \eqref{eqdep} soient conditionnées sur $\bar{I}$ ou encore sur $\{P\cup \bar{P}\}$.
Afin de pouvoir combiner les fonctions de masse \eqref{eq35} et \eqref{eqdep} pour tenir compte du degré de dépendance des clusters dans la fonction de masse de la dépendance positive ou négative, il faut déconditionner les fonctions de masse conditionnelles et redéfinir les deux fonctions de masse sur un cadre le discernement commun $\Pos$.
Le déconditionnement de la fonction de masse conditionnelle de l'équation \eqref{eqdep} utilisant l'équation \eqref{decon} permet de retrouver la fonction de masse la moins informative en supprimant l'hypothèse forte sur la dépendance de tous les clusters liés. Les fonctions de masse obtenues sont alors~:
\begin{equation}
\label{depraff}
\left\{
\begin{tabular}{ll}
%$m^{\Pos}_{{k_1},{k_2}}(\emptyset)=0$\\
$m^{\Pos}_{{k_i},{k_j}}(P\cup I)=1-Conf(Cl^i_{k_i},Cl^j_{k_j})$\\
$m^{\Pos}_{{k_i},{k_j}}(\bar{P} \cup I )=Conf(Cl^i_{k_i},Cl^j_{k_j})$\\
%$m^{\Pos}_{{k_i},{k_j}}(P\cup \bar{P} \cup I)=0$\\
\end{tabular}
\right.
\end{equation}

Le cadre de discernement $\I=\{\bar{I},I\}$ peut être raffiné en raffinant l'hypothèse $\bar{I}=\{P\cup \bar{P}\}$, ceci mènera au cadre de discernement raffiné $\Pos$. Les fonctions de masse marginales de la dépendance des clusters liés de l'équation~\eqref{eq35} deviennent après raffinement~:
\begin{equation}
\label{raff}
\left\{
\begin{tabular}{ll}
%$\displaystyle{m^{\Omega_I}_{{k_1},{k_2}}}(\emptyset)=0$\\
$m^{\Pos}_{{k_i},{k_j}}(I)=\alpha^i_{{k_i},{k_j}} \, (1-\beta^i_{{k_i},{k_j}})$\\
$m^{\Pos}_{{k_i},{k_j}}(P \cup \bar{P})=\alpha^i_{{k_i},{k_j}} \, \beta^i_{{k_i},{k_j}}$\\
$m^{\Pos}_{{k_i},{k_j}}(I\cup P \cup \bar{P})=1-\alpha^i_{{k_i},{k_j}}$\\
\end{tabular}
\right.
\end{equation}
Nous définissons ainsi la fonction de masse de l'indépendance, dépendance positive et dépendance négative de chaque couple de clusters liés de $S_1$ et $S_2$ après combinaison conjonctive des fonctions de masse des équations \eqref{depraff} et \eqref{raff} définies sur le cadre de discernement $\Pos$~:
\begin{equation}
\label{m_indep}
\left\{
\begin{tabular}{ll}
%$m^{\Omega_P}_{{k_1},{k_2}}(\emptyset)=0$\\
$m^{\Pos}_{{k_i},{k_j}}(I)=\alpha^{i}_{k_i,k_j} \, (1-\beta^i_{{k_i},{k_j}})$\\
$m^{\Pos}_{{k_i},{k_j}}(P)=\alpha^{i}_{k_i,k_j} \, \beta^i_{{k_i},{k_j}}(1-Conf(Cl^i_{k_i},Cl^j_{k_j}))$\\
$m^{\Pos}_{{k_i},{k_j}}(\bar{P})=\alpha^{i}_{k_i,k_j} \, \beta^i_{{k_i},{k_j}} Conf(Cl^i_{k_i},Cl^j_{k_j})$\\
$m^{\Pos}_{{k_i},{k_j}}(I\cup P)=(1-\alpha^{i}_{k_i,k_j}) \, (1-Conf(Cl^i_{k_i},Cl^j_{k_j})) $\\
$m^{\Pos}_{{k_i},{k_j}}(I\cup \bar{P})=(1-\alpha^{i}_{k_i,k_j}) \, Conf(Cl^i_{k_i},Cl^j_{k_j})$\\
\end{tabular}
\right.
\end{equation}

La fonction de masse générale sur la dépendance de la source $S_1$ par rapport à $S_2$ est donnée par~:
\begin{equation}
\label{ind2}
m^{\Pos} (A)=\frac{1}{C}\displaystyle{\sum_{k_i=1}^C} m^{\Pos}_{{k_i},{k_j}}(A)
\end{equation} 
avec $\{i,j\}\in\{1,2\}$ et $i\neq j$, où $k_2$ est le cluster de la source $S_2$ associé au cluster $k_1$ de la source $S_1$. Cette fonction de masse représente ainsi l'ensemble des croyances élémentaires sur l'indépendance, la dépendance positive et négative de la source $S_1$ face à la source $S_2$. Cette fonction de masse est la combinaison avec la moyenne de toutes les fonctions de masse sur l'indépendance, dépendance positive et dépendance négative de tous les clusters liés.

Notons aussi qu'une source est positivement dépendante d'elle-même puisqu'en appliquant l'algorithme de classification nous obtenons exactement la même répartition de classes ce qui impliquera $I_d(S,S)=0$ et $m^{\Pos}(P)=1$.
Le degré d'indépendance est la probabilité pignistique de l'hypothèse $I$, $BetP(I)$, ceux des dépendances positive et négative sont respectivement $BetP(P)$ et $BetP(\bar{P})$.
%%%%%%%%%%%%%%%%%%%%%%%%%%%%%%%%%%%%%%%%%%%%%%%%%%%%%%%%%%%%%%%%%%%%%%%%%%%%%%%%%%%%%%%%%%%%%%%%%%%%%%%%%%%%%%%%%%%%%%%%%%%%%%%%
\section{Intégration de l'indépendance dans une fonction de masse}
\label{IntégrationInd}
%%%%%%%%%%%%%%%%%%%%%%%%%%%%%%%%%%%%%%%%%%%%%%%%%%%%%%%%%%%%%%%%%%%%%%%%%%%%%%%%%%%%%%%%%%%%%%%%%%%%%%%%%%%%%%%%%%%%%%%%%%%%%%
Nous avons vu que l'indépendance est généralement une information supplémentaire nécessaire à la fusion d'informations, mais non prise en compte dans le formalisme choisi. La section~\ref{independance} propose une modélisation et estimation d'une mesure d'indépendance dans le cadre de la théorie des fonctions de croyance. Nous allons ici nous appuyer sur le principe de l'affaiblissement présenté dans la section~\ref{affaiblissement} afin de tenir compte de l'indépendance dans les fonctions de masse en vue de la combinaison. 

En effet, lors de la combinaison conjonctive par exemple, l'hypothèse d'indépendance cognitive des sources d'informations est nécessaire. Si les sources sont dépendantes on peut penser qu'elles ne devraient pas être combinées par ce biais. Cependant, comme le montre la section~\ref{independance} les sources peuvent avoir des degrés de dépendance et d'indépendance. L'information fournie sur l'indépendance n'est pas catégorique. Ainsi, combiner deux sources fortement indépendantes devraient revenir à la combinaison de deux sources indépendantes. 
Si une source est dépendante d'une autre source, nous pouvons considérer que cette première source ne doit pas influer la combinaison avec la seconde. Ainsi cette source doit représenter l'élément neutre de la combinaison. 

Dans ce cas, il suffit d'appliquer la procédure d'affaiblissement de la section~\ref{affaiblissement} sur la fonction de masse $m^\Omega$ de la source $S_1$ en considérant l'indépendance donnée par la fonction de masse de l'équation~\eqref{ind2} au lieu de celle de l'équation~\eqref{fiab} dans le cas de la fiabilité.

\`A présent, nous distinguons la dépendance positive de la dépendance négative. Si une source est dépendante positivement d'une autre source, il ne faut pas en tenir compte et donc tendre vers un résultat de combinaison qui prendrait cette première source comme un élément neutre. Enfin si une source est dépendante négativement d'une autre source, il peut être intéressant de marquer cette dépendance conflictuelle en augmentant la masse sur l'ensemble vide.

Pour réaliser ce schéma, nous proposons d'affaiblir les fonctions de masse d'une source $S_1$ en fonction de sa mesure d'indépendance à une autre source $S_2$, donnée par la fonction de masse $m_{i}^{\I}$ de l'équation~\eqref{ind2}.

Nous considérons ici une fonction de masse d'une source $m^\Omega$ en fonction de son indépendance ou dépendance à une autre source. Ainsi nous définissons~:
\begin{equation}
\label{m_indep}
\left\{
\begin{tabular}{ll}
%$m^{\Omega_P}_{{k_1}{k_2}}(\emptyset)=0$\\
$m^\Omega{[I]}(X)=m^\Omega(X)$\\
$m^\Omega{[\bar{P}]}(X)=m^\Omega(X)$&$m^\Omega(X)=1\text{ si }X=\emptyset\text{, }0\text{ sinon}$\\
$m^\Omega{[P]}(X)=m^\Omega(X)$&$m^\Omega(X)=1\text{ si }X=\Omega\text{, }0\text{ sinon}$\\
\end{tabular}
\right.
\end{equation}
Suivant la procédure d'affaiblissement, nous effectuons une extension à vide sur la fonction de masse $m^{\I}$~:
\begin{equation}
m^{\I\uparrow\Omega\times\I} \left(Y\right)=\left\lbrace
\begin{array}{ll}
 m^{\I} \left(X\right)&\text{si } Y=\Omega \times X, \quad X\subseteq\I\\
0 &\text{sinon}
\end{array}
\right.
\end{equation}
Le déconditionnement des fonctions de masse $m^\Omega[I]$, $m^\Omega{[P]}$ et $m^\Omega{[\bar{P}]}$ est donné par~:
\begin{equation}
m^{\Omega\Uparrow\Omega\times\I}{[I]}((A\times I)\cup(\Omega\times\overline{I}))=m^\Omega{[I]}(A), \quad A\subseteq\Omega
\end{equation}
où $\bar{I}=\{P\cup \bar{P}\}$ est un raffinement.
\begin{equation}
m^{\Omega\Uparrow\Omega\times\I}{[\bar{P}]}((A\times \bar{P})\cup(\Omega\times \{I\cup P\}))=m^\Omega{[\bar{P}]}(A), \quad A\subseteq\Omega
\end{equation}
\begin{equation}
m^{\Omega\Uparrow\Omega\times\I}{[P]}((A\times P)\cup(\Omega\times \{I\cup\bar{P}\}))=m^\Omega{[P]}(A), \quad A\subseteq\Omega
\end{equation}
Ce dernier déconditionnement mène en fait à la masse de l'ignorance et est l'élément neutre de la combinaison conjonctive. 

Nous réalisons ensuite la combinaison conjonctive~:
\begin{equation}
m_\conj^{\Omega\times\I}(Y)= m^{\I\uparrow\Omega\times\I} \ocap m^{\Omega\Uparrow\Omega\times\I}{\left[I\right]} \ocap m^{\Omega\Uparrow\Omega\times\I}{\left[\bar{P}\right]} (Y), \quad \forall Y\subset \Omega\times\I
\end{equation}

La marginalisation de la fonction de masse permet ensuite de revenir dans l'espace $\Omega$~:
\begin{equation}
 m^{\Omega\times\I\downarrow\Omega}
\left(X\right)=\displaystyle{
\sum_{
\left\lbrace
Y\subseteq\Omega\times\I | Proj\left(Y\downarrow\Omega\right)=X
\right\rbrace}
m_\conj^{\Omega\times\I}\left(Y\right)
}
\end{equation}

Cette procédure réalisée pour la source $S_1$ en rapport à la source $S_2$ peut être réalisée pour la source $S_2$ au regard de la source $S_1$. Ainsi les deux fonctions de masse obtenues peuvent être combinées par la règle de combinaison conjonctive qui suppose l'indépendance. 
%%%%%%%%%%%%%%%%%%%%%%%%%%%%%%%%%%%%%%%%%%%%%%%%%%%%%%%%%%%%%%%%%%%%%%%%%%%%%%%%%%%%%%%%%%%%%%%%%%%%%%%%%%%%%%%%%%%%%%%%%%%%%%%%
\section{Expérimentations}
\label{exp}
%%%%%%%%%%%%%%%%%%%%%%%%%%%%%%%%%%%%%%%%%%%%%%%%%%%%%%%%%%%%%%%%%%%%%%%%%%%%%%%%%%%%%%%%%%%%%%%%%%%%%%%%%%%%%%%%%%%%%%%%%%%%%%
Pour tester la méthode précédemment décrite, nous avons généré des fonctions de masse aléatoirement. Tous les tirages aléatoires sont fait suivant la loi uniforme. L'algorithme~\ref{Mass generating} est utilisé pour générer $n$ fonctions de masse.
\begin{algorithm}[h]
\caption{Générer $n$ fonctions de masse}
\label{Mass generating}
\begin{algorithmic}[1]
\REQUIRE $|\Omega|$, $n:$ nombre des fonctions de masse à générer
\FOR {$i=1$ to $n$}
   \STATE Tirer aléatoirement $\mid F\mid$, le nombre d'éléments focaux dans l'intervalle  $[1,| 2^\Omega |]$.
	 \STATE Tirer aléatoirement $\mid F\mid$ éléments focaux notés $F$.
	 \STATE Couper l'intervalle $[0,1]$ en $\mid F\mid-1$ sous-intervalles aléatoires continus.
		\FOR {$j=1$ to $\mid F \mid$}
			\STATE La masse de chaque élément focal est la longueur de l'un des sous-intervalles.
		\ENDFOR
\ENDFOR
\RETURN $n$ fonctions de masse.
\end{algorithmic}
\end{algorithm}

Cet algorithme a été utilisé pour générer $100$ fonctions de masse, définies sur un cadre de discernement de cardinalité $5$, pour deux sources $S_1$ et $S_2$ avec les trois hypothèses sur la dépendance des sources (sources indépendantes, sources dépendantes positivement et sources dépendantes négativement), les résultats des tests sont présentés dans le tableau \ref{tab}.
\begin{enumerate}
\item{Sources indépendantes~:} Supposons que deux sources $S_1$ et $S_2$ sont complètement indépendantes. Nous avons alors généré $100$ fonctions de masse pour chaque source comme décrit dans l'algorithme \ref{Mass generating} avec $\mid\Omega\mid=5$. %AM Les fonctions de masse générées ne sont pas modifiées. 

\item{Sources dépendantes positivement:} Lorsque deux sources $S_1$ et $S_2$ sont dépendantes positivement, les classes de décision (en terme de probabilité pignistique), calculées à partir des fonctions de masse qu'elles fournissent, sont les mêmes. Les deux sources $S_1$ et $S_2$ s'expriment de la même manière puisqu'elles sont dépendantes. Nous avons généré aléatoirement $100$ fonctions de masse pour chaque source. Ces fonctions de masse sont modifiées par la suite suivant l'algorithme~\ref{Positive mass generating}.
\begin{algorithm}[h]
\caption{Générer des fonctions de masse dépendantes positivement}
\label{Positive mass generating}
\begin{algorithmic}[1]
\REQUIRE $n$ fonctions de masse générées aléatoirement avec l'algorithme \ref{Mass generating}, Les classes de décision.
\FOR {$i=1$ to $n$}
\STATE Recherche des éléments focaux $F$ de chaque fonction de masse $m_i$
  \FOR {$j=1$ to $\mid F\mid$}
  \STATE La masse affectée au $j^{\text{ème}}$ élément focal est transférée à son union avec la classe de décision.  
\ENDFOR
\ENDFOR
\RETURN $n$ fonctions de masse modifiées.
\end{algorithmic}
\end{algorithm}

\item{Sources dépendantes négativement~:} Lorsque deux sources $S_1$ et $S_2$ sont dépendantes négativement, les classes de décision (en terme de probabilité pignistique), calculées à partir des fonctions de masse qu'elles fournissent, sont contradictoires mais d'une façon ordonnée. Les deux sources $S_1$ et $S_2$ sont dépendantes mais l'une des deux sources a tendance à dire l'opposé de l'autre. Nous avons généré aléatoirement $100$ fonctions de masse pour chaque source. Ces fonctions de masse sont modifiées par la suite suivant l'algorithme \ref{Negative mass generating}. 
\begin{algorithm}[h]
\caption{Générer des fonctions de masse dépendantes négativement}
\label{Negative mass generating}
\begin{algorithmic}[1]
\REQUIRE $n$ fonctions de masse générées aléatoirement avec l'algorithme \ref{Mass generating} pour une source, Les classes de décision de l'autre source, La correspondance de classes contradictoires.
\FOR {$i=1$ to $n$}
\STATE Recherche des éléments focaux $F$ de chaque fonction de masse $m_i$.
  \FOR {$j=1$ to $\mid F\mid$}
  \STATE Transférer la masse de l'élément focal $j$ à son union avec la classe contradictoire (la classe contradictoire à la classe de décision de $m_i$) privé de la classe de décision de $m_i$.  
\ENDFOR
\ENDFOR
\RETURN $n$ fonctions de masse modifiées.
\end{algorithmic}
\end{algorithm}

\end{enumerate}
\begin{table}[ht]
\begin{center}
\begin{tabular}{|l|l|l|l|}
\hline
\scriptsize{\bfseries Type de dépendance} & \scriptsize{\bfseries Degrés d'indépendance, dépendance positive et dépendance négative} \\
\hline
\multirow{2}{2cm}{Indépendance}
&\\
 &$I_d(S_1,S_2)=0.72$, $\bar{I_d}(S_1,S_2)=0.28$  \\  
 &$I_d(S_2,S_1)=0.66$, $\bar{I_d}(S_2,S_1)=0.34$ \\ 
\hline 
\multirow{2}{2cm}{Dépendance positive}
&\\
 &$m^{\I,1}(I)=0.26$, $m^{\I,1}(P)=0.56,m^{\I,1}(\bar{P})=0.18$\\
 &$m^{\I,2}(I)=0.35$, $m^{\I,2}(P)=0.5,m^{\I,2}(\bar{P})=0.15$ \\    
\hline
\multirow{2}{2cm}{Dépendance négative}
&\\
 &$m^{\I,1}(I)=0.35$, $m^{\I,1}(P)=0.25,m^{\I,1}(\bar{P})=0.4$\\
 &$m^{\I,2}(I)=0.38$, $m^{\I,2}(P)=0.18,m^{\I,2}(\bar{P})=0.44$ \\   
\hline
\end{tabular}
\caption{Résultats des tests sur $100$ fonctions de masse générées}
 \label{tab}
 \end{center}
\end{table}
La complexité temporelle\footnote{Les expérimentations ainsi que la complexité temporelle ont été testé sous Matlab R2010a.} de l'algorithme proposé dépend fortement de la cardinalité du cadre de discernement. Dans le cas illustré, $\mid \Omega\mid=5$, la complexité est de quelques secondes mais elle peut être beaucoup plus importante pour les plus grands cadres de discernement. Notons que la complexité temporelle de l'algorithme de classification est optimisée puisque les distances entre les fonctions de masse ne sont calculées qu'une seule fois.
%%%%%%%%%%%%%%%%%%%%%%%%%%%%%%%%%%%%%%%%%%%%%%%%%%%%%%%%%%%%%%%%%%%%%%%%%%%%%%%%%%%%%%%%%%%%%%%%%%%%%%%%%%%%%%%%%%%%%%%%%%%%%%%%%%%%%%%%%%%%%%%%%%%%%%%%%%%%%%%%%%%%%%%%%%%%%%%%%
\subsection{Fonctionnement de l'affaiblissement par la mesure d'indépendance}
%%%%%%%%%%%%%%%%%%%%%%%%%%%%%%%%%%%%%%%%%%%%%%%%%%%%%%%%%%%%%%%%%%%%%%%%%%%%%%%%%%%%%%%%%%%%%%%%%%%%%%%%%%%%%%%%%%%%%%%%%%%%%%%%%%%%%%%%%%%%%%%%%%%%%%%%%%%%%%%%%%%%%%%%%%%%%%%%%
Nous allons, dans un premier temps, illustrer le fonctionnement de l'affaiblissement par la mesure d'indépendance. Nous considérons ici un cadre de discernement $\Omega=\{\omega_1,\omega_2,\omega_3\}$. Supposons que nous ayons deux sources $S_1$ et $S_2$ donnant deux fonctions de masse~:
\begin{eqnarray}
\label{massesources1}
m_1^\Omega(\omega_1)=0.2, \, m_1^\Omega(\omega_1 \cup \omega_2)=0.5, \, m_1^\Omega(\Omega)=0.3,\\
\label{massesources2}
m_2^\Omega(\omega_2)=0.1, \, m_2^\Omega(\omega_1 \cup \omega_2)=0.6, \, m_2^\Omega(\Omega)=0.3
\end{eqnarray}
La combinaison conjonctive donne~:
\begin{eqnarray*}
&& m_{1\ocap2}^\Omega(\emptyset)=0.02, \, m_{1\ocap2}^\Omega(\omega_1)=0.18,\, m_{1\ocap2}^\Omega(\omega_2)=0.08,\\ 
&& m_{1\ocap2}^\Omega(\omega_1 \cup \omega_2)=0.63, \, m_{1\ocap2}^\Omega(\Omega)=0.09
\end{eqnarray*}

Cette combinaison conjonctive est effectuée avec l'hypothèse d'indépendance cognitive des deux sources. Si une connaissance externe permet de mesurer la dépendance positive et négative de la source $S_1$ par rapport à la source $S_2$ telle que fournie par l'équation~\eqref{eq35}, nous devons en tenir compte avant la combinaison conjonctive. \\
Supposons une fonction de masse traduisant donc une forte dépendance positive de $S_1$ par rapport à $S_2$. Nous avons ainsi la fonction de masse suivante~:
\begin{equation}
\left\{
\begin{tabular}{ll}
$m^{\Pos}(I)= 0.26$\\
$m^{\Pos}(P)=0.56$\\
$m^{\Pos}(\bar{P})=0.18$\\
\end{tabular}
\right.
\end{equation}
Notons que $m^{\Pos}(I \cup P)=0$ et $m^{\Pos}(I \cup \bar{P})=0$ puisque les facteurs d'affaiblissement $\alpha^{i}$ ne sont pas définis donc fixés à $1$.
Le tableau~\ref{details} présente les différentes étapes d'extension à vide, de déconditionnement et de combinaison dans l'espace $\Omega\times\Pos$. L'extension à vide et le déconditionnement transfèrent les masses sur les éléments focaux correspondant de l'espace $\Omega\times\Pos$. La combinaison des trois fonctions de masse dans cet espace fait apparaître la masse sur l'ensemble vide qui correspond à la part de dépendance négative.
 % Modifier le 31/06 Mouloud
\begin{table}[ht]
\begin{center}
\begin{tabular}{|c|c|c|c|c|}
\hline
 & & & & \\
focal & $m^{\Pos\uparrow\Omega\times\Pos}$ & $m^\Omega[I]^{\Uparrow\Omega\times\Pos}$ & \!\!$m^\Omega[\bar{P}]^{\Uparrow\Omega\times\Pos}$\!\! & $m_\conj^{\Omega\times\Pos}$\\
\hline
%\vspace{-0.3cm} 				&  	& 	& 	&\\
$\emptyset$ 					& 	& 	&  	&0.18 \\
\hline
%\vspace{-0.3cm} 				&  	& 	& 	&\\
$\omega_1 \times I$	 			&  	& 	& 	&0.052 \\
\hline
%\vspace{-0.3cm} 				&  	& 	& 	&\\
$(\omega_1 \cup \omega_2) \times I$ 		&  	& 	& 	&0.13 \\
\hline
%\vspace{-0.3cm} 				&  	&	& 	&\\
$\Omega \times I$ 				& 0.26	& 	& 	&0.078 \\
\hline
%\vspace{-0.3cm} 				&  	& 	& 	&\\
$\Omega \times P$ 				& 0.56 	& 	& 	&0.56 \\
\hline
%\vspace{-0.3cm} 				&  	& 	& 	&\\
$(\omega_1 \times I) \cup (\Omega \times P) $ 	&  	& 	& 	&\\
\hline
%\vspace{-0.3cm} 				&  	& 	& 	&\\
$((\omega_1 \cup \omega_2) \times I) \cup (\Omega \times P)$ &  	& 	& 	& \\
\hline
%\vspace{-0.3cm} 				&  	& 	& 	&\\
$\Omega \times \bar{P}$ 			& 0.18 	& 	& 	&\\
\hline
%\vspace{-0.3cm} 				&  	& 	& 	&\\
$\Omega \times (I \cup  P)$ 		&  	& 	& 1	&\\
\hline
%\vspace{-0.3cm} 				&  	& 	& 	&\\
$(\omega_1 \times I) \cup (\Omega \times (P \cup  \bar{P}))$ 			&  	& 0.2	& 	&\\
\hline
%\vspace{-0.3cm} &  & & &\\
$((\omega_1 \cup \omega_2) \times I) \cup (\Omega \times (P \cup  \bar{P}))$	 &  	&0.5 	& 	&\\
\hline
%\vspace{-0.3cm} &  & & &\\
$\Omega \times \Pos$ 			&  & 0.3 & & \\
\hline 
\end{tabular}
\caption{Détails de l'affaiblissement de la mesure d'indépendance~: fonctions de masse dans $\Omega\times\Pos$.}
\label{details}

\end{center}
\end{table}

Le tableau~\ref{details2} présente ensuite la marginalisation et le résultat de combinaison avec la fonction de masse $m^\Omega_2$ non modifiée ({\em i.e.} que l'hypothèse d'indépendance totale de $S_2$ par rapport à $S_1$ est faite). Nous constatons que la masse transférée sur l'ignorance ne devient plus importante que lors de la combinaison conjonctive sans hypothèse sur la dépendance positive.
% Modifier le 31/06 Mouloud
\begin{table}[ht]
\begin{center}
\begin{tabular}{|c|c|c|c|}
\hline
focal & $m_1^{\Omega\times\Pos \downarrow \Omega}$ & $m_2^{\Omega}$ & $m_1^{\Omega\times\Pos \downarrow \Omega} \ocap m_2^{\Omega}$\\
\hline
$\emptyset$ 		& 0.18 & & 0.25432\\
\hline
$\omega_1$ 		& 0.052 &  & 0.0468 \\
\hline
$\omega_2$ 		&	& 0.1 & 0.00768 \\
\hline
$\omega_1 \cup \omega_2$ &  0.13 & 0.6 & 0.15528 \\
\hline
$\Omega$ 		&0.638  & 0.3 & 0.53592 \\
\hline 
\end{tabular}
\caption{Détails de l'affaiblissement de la mesure d'indépendance~: marginalisation et combinaison}
\label{details2}
\end{center}
\end{table}

\begin{figure}[ht]
\begin{center}
\includegraphics[width=0.7\textwidth, keepaspectratio]{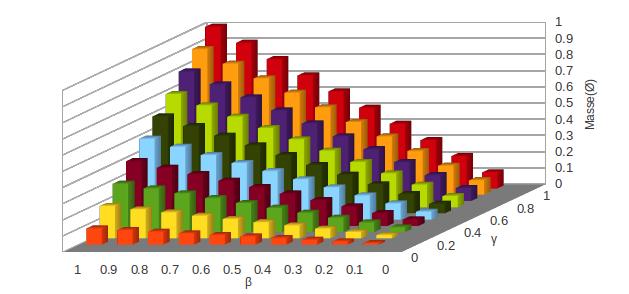}
\end{center}
\caption{Variation de la masse sur l'ensemble vide pour un affaiblissement par la mesure d'indépendance d'une masse dogmatique.}
\label{fig1}
\end{figure}

\begin{figure}[ht]
\begin{center}
\includegraphics[width=0.7\textwidth, keepaspectratio]{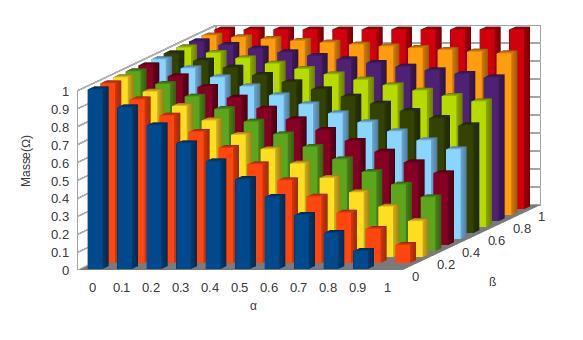}
\end{center}
\caption{Variation de la masse sur l'ignorance $\Omega$ pour un affaiblissement par la mesure d'indépendance d'une masse dogmatique.}
\label{fig2}
\end{figure}

Afin de bien illustrer le transfert de masse sur l'ensemble vide et sur l'ignorance, les figures~\ref{fig1} et \ref{fig2} représentent les masses en fonction des variations de $\alpha_i$ (représentant un facteur d'affaiblissement de la source $S_i$) , $\beta_i$ (représentant le taux de dépendance de $S_i$ face à $S_j$) et $\gamma_i$ (représentant le taux de dépendance négative de $S_i$ face à $S_j$) d'une fonction de masse décrite par~\eqref{m_indep} pour une fonction de masse dogmatique quelconque de la source $S_i$. Ainsi sur la figure~\ref{fig1}, représentant les variations de masse sur l'ensemble vide, $\alpha_i$ est fixé à 1, $\beta_i$ et $\gamma_i$ varient. Sur la figure~\ref{fig2}, représentant les variations de masse sur l'ignorance, $\gamma_i$ est fixé à 1, $\alpha_i$ et $\beta_i$ varient.

La figure~\ref{fig1} montre ainsi que, pour la source $S_i$, plus $\beta_i$ et $\gamma_i$ sont grands plus on obtient une masse importante sur l'ensemble vide et donc une dépendance négative. La quantité $\beta_i$ représente la part de dépendance de la source et la quantité $\gamma_i$ représente la part de dépendance négative.

La figure~\ref{fig2} présente quand à elle, la variation de la masse sur $\Omega$, l'ignorance. Cette masse est donnée directement par $\alpha_i (1 - \beta_i)$ qui contient donc la part d'indépendance $(1 - \beta_i)$ et la fiabilité $\alpha_i$ de la source $S_i$. 

Nous illustrons ainsi le résultat escompté de l'affaiblissement par la mesure d'indépendance, c'est-à-dire que nous retrouvons sur la masse de l'ensemble vide la quantité de dépendance négative (associé à la part de dépendance de la source) et sur l'ignorance la quantité de fiabilité et d'indépendance. 
Nous remarquons aussi que lorsque la source est fiable ($\alpha_i=1$) et indépendante ($\beta_i=0$), la fonction de masse de la source n'est pas modifiée.

\subsection{Influence sur le résultat de combinaison}
Afin d'illustrer l'influence de la prise en compte de la mesure d'indépendance sur les fonctions de masse, nous allons considérer ici les deux sources précédentes $S_i$ et $S_j$ qui fournissent les fonctions de masse données par les équations~\eqref{massesources1} et \eqref{massesources2}. Nous allons considérer trois cas pour chaque source avec un cas où la source $S_i$ est plutôt indépendante de $S_j$ ($\alpha_i=0.95$, $\beta_i=0.95$, $\gamma_i=0.05$), un cas où elle est plutôt dépendante positivement ($\alpha_i=0.95$, $\beta_i=0.05$, $\gamma_i=0.95$) et un cas où elle est plutôt dépendante négativement ($\alpha_i=0.95$, $\beta_i=0.05$, $\gamma_i=0.05$). Pour la source $S_j$, nous considérons trois cas moins catégoriques en fixant la fiabilité $\alpha_j=0.9$~: le cas plutôt indépendant ($\beta_j=0.9$, $\gamma_j=0.1$), le cas plutôt dépendant positivement ($\beta_j=0.1$, $\gamma_j=0.9$) et le cas plutôt dépendant négativement ($\beta_j=0.1$, $\gamma_j=0.1$).

\begin{table}[ht]
\begin{center}
\begin{tabular}{|c|c|c|c|c|c|c|c|c|}
\hline
 & & & \multicolumn{6}{c|} {$S_j$ : $\alpha_j=0.9$} \\
\cline{4-9}
cas & élément &  & \multicolumn{2}{c|}{$\beta_j=0.9$, $\gamma_j=0.1$} & \multicolumn{2}{c|}{$\beta_j=0.1$, $\gamma_j=0.9$} & \multicolumn{2}{c|}{$\beta_j=0.1$, $\gamma_j=0.1$}\\
%Mouloud
\cline{4-9}
%\vspace{-0.3cm} &  & & & & & & & \\
 & focal & \!\!\!\! $m_1^{\Omega\times\Pos\downarrow\Omega}$\!\!\!\! &\!\!\!\!  $m_2^{\Omega\times\Pos\downarrow\Omega}$ \!\!\!\!&  \!\!\!\!$m_{1\cap2}^{\Omega\times\Pos\downarrow\Omega}$\!\!\!\! & \!\!\!\!  $m_2^{\Omega\times\Pos\downarrow\Omega}$ \!\!\!\!& \!\!\!\! $m_{1\cap2}^{\Omega\times\Pos\downarrow\Omega}$\!\!\!\! & \!\!\!\! $m_2^{\Omega\times\Pos\downarrow\Omega}$\!\!\!\! &\!\!\!\!  $m_{1\cap2}^{\Omega\times\Pos\downarrow\Omega}$\!\!\!\! \\
  %  \cline{2-3} 
\hline
$S_i$ & $\emptyset$                         & 0.045125 & 0.081 & 0.12257 &  0.081 & 0.12337    &0.009 &0.0545389 \\
\cline{2-9}
$\alpha_i=0.95$ & $\omega_1$                 & 0.01 &        & 0.00909    &          & 0.00829 &      &0.00909 \\
\cline{2-9}
 $\beta_i=0.95$ & $\omega_2$                 &        &  0.01 & 0.0779175   & 0.09 & 0.0850388  &0.082 &0.0774798 \\
\cline{2-9}
 $\gamma_i=0.05$ & $\omega_1 \cup \omega_2$  & 0.025  & 0.06   & 0.07138     &0.54 & 0.517457 &0.492 & 0.475303 \\
\cline{2-9}
 & $\Omega$                                & 0.919875 & 0.849 & 0.780974     & 0.289 & 0.265844 & 0.417 & 0.383588 \\
\hline
\hline
$S_i$ & $\emptyset$                       & 0.045125 & 0.081 & 0.12437   & 0.081 & 0.13957 &0.009 & 0.0692989 \\
\cline{2-9}
$\alpha_i=0.95$ & $\omega_1$                & 0.19 &         & 0.17271   &       & 0.15751 &      & 0.17271 \\
\cline{2-9}
 $\beta_i=0.05$ & $\omega_2$                &        &0.01    & 0.00764875 & 0.09 &0.0688388 & 0.082 & 0.0627198 \\
\cline{2-9}
 $\gamma_i=0.95$ & $\omega_1 \cup \omega_2$ & 0.475 &0.06   & 0.449167   &0.54 & 0.550307 & 0.492 & 0.574394 \\
\cline{2-9}
 & $\Omega$                              & 0.289875 & 0.849  & 0.246104   & 0.289 & 0.0837739 & 0.417 &0.120878 \\
\hline
\hline
$S_i$ & $\emptyset$                      & 0.002375 & 0.081 & 0.0849926    & 0.081& 0.0994726 &0.009 &0.0261956 \\
\cline{2-9}
$\alpha_i=0.95$ & $\omega_1$               & 0.181   &         & 0.164529   &       & 0.150049 &     & 0.164529 \\
\cline{2-9}
 $\beta_i=0.05$ & $\omega_2$               &          &0.01    & 0.00816625 & 0.09& 0.0734962 &0.082 & 0.0669633 \\
\cline{2-9}
 $\gamma_i=0.05$ & $\omega_1 \cup \omega_2$ & 0.4525  &0.06   & 0.43317   & 0.54 & 0.57175 & 0.492 & 0.590472 \\
\cline{2-9}
 & $\Omega$                              & 0.364125 & 0.849   & 0.309142   &0.289 & 0.105232 & 0.417 & 0.15184 \\
\hline
\end{tabular}
\caption{Résultats de combinaison selon les hypothèses de dépendance et d'indépendance des deux sources $S_i$ et $S_j$.}
\label{res_comb}

\end{center}
\end{table}

Ainsi, le tableau~\ref{res_comb} présente les résultats de la combinaison des deux sources en fonction des hypothèses d'indépendance et de dépendance positive ou négative des deux sources $S_1$ et $S_2$. Nous constatons que lorsque les deux sources sont plutôt indépendantes, les résultats obtenus sont proches de la combinaison conjonctive sous l'hypothèse d'indépendance. Lorsqu'une des deux sources est dépendante négativement de l'autre, la masse transférée sur l'ensemble vide est importante. 
Lorsque l'une des deux sources est dépendante positivement de l'autre, la masse est transférée sur l'ignorance mais de façon moins importante que pour la dépendance négative. En effet, l'ensemble vide est un élément absorbant pour la combinaison conjonctive. Cette masse sur l'ensemble vide, à l'issue de la combinaison conjonctive, peut ainsi jouer un rôle d'alerte sur la dépendance négative. Une autre alternative serait d'envisager une autre règle de combinaison lorsque la masse issue de la 
dépendance négative est très importante.

%%%%%%%%%%%%%%%%%%%%%%%%%%%%%%%%%%%%%%%%%%%%%%%%%%%%%%%%%%%%%%%%%%%%%%%%%%%%%%%%%%%%%%%%%%%%%%%%%%%%%%%%%%%%%%%%%%%%%%%%%%%%%%%%
\section{Conclusion et perspectives}
%%%%%%%%%%%%%%%%%%%%%%%%%%%%%%%%%%%%%%%%%%%%%%%%%%%%%%%%%%%%%%%%%%%%%%%%%%%%%%%%%%%%%%%%%%%%%%%%%%%%%%%%%%%%%%%%%%%%%%%%%%%%%%
Dans cet article, nous avons proposé une méthode d'apprentissage de l'indépendance de sources afin de justifier l'hypothèse sur l'indépendance lors du choix des règles de combinaison à utiliser pour la fusion. Nous avons également proposé d'estimer la dépendance positive et négative afin de pouvoir tenir compte de cette information dans les fonctions de masse avant de les combiner. Cette nouvelle information sur la dépendance des sources peut être intégrée dans les fonctions de masse fournies par ces sources avant de les combiner. Une autre solution à la dépendance des sources est de proposer une règle de combinaison tenant compte de l'indépendance, dépendance positive et négative des sources. Dans les prochains travaux, nous proposerons une règle de combinaison mixant la combinaison disjonctive et la combinaison prudente en fonction des degrés de dépendance. Nous définirons également le facteur d'affaiblissement $\alpha^i_{k_i,k_j}$ permettant de tenir compte du nombre de fonctions de masse dans les clusters appariés lors de l'apprentissage de leurs indépendances.
%%%%%%%%%%%%%%%%%%%%%%%%%%%%%%%%%%%%%%%%%%%%%%%%%%%%%%%%%%%%%%%%%%%%%%%%%%%%%%%%%%%%%%%%%%%%%%%%%%%%%%%%%%%%%%%%%%%%%%%%%%%%%%%
\bibliographystyle{rnti}
\bibliography{biblio}

\providecommand\Fr{}
\providecommand\Eng{}
\providecommand\andname{and}
\providecommand\andnamec{and}

\begin{thebibliography}{}


\bibitem[{Ben{ }Hariz et~al.}(2006){Ben{ }Hariz, Elouedi, \andnamec{}
  Mellouli}]{BenHariz06}
Ben{ }Hariz, S., Z.~Elouedi, \andname{} K.~Mellouli (2006).
\newblock Clustering approach using belief function theory.
\newblock In {\em AIMSA}, pp.\  162--171.

\bibitem[{Ben{ }Yaghlane et~al.}(2002a){Ben{ }Yaghlane, Smets, \andnamec{}
  Mellouli}]{BenYaghlane02a}
Ben{ }Yaghlane, B., P.~Smets, \andname{} K.~Mellouli (2002a).
\newblock Belief function independence: I. the marginal case.
\newblock {\em Int. J. Approx. Reasoning\/}~{\em 29\/}(1), 47--70.

\bibitem[{Ben{ }Yaghlane et~al.}(2002b){Ben{ }Yaghlane, Smets, \andnamec{}
  Mellouli}]{BenYaghlane02b}
Ben{ }Yaghlane, B., P.~Smets, \andname{} K.~Mellouli (2002b).
\newblock Belief function independence: Ii. the conditional case.
\newblock {\em Int. J. Approx. Reasoning\/}~{\em 31\/}(1-2), 31--75.

\bibitem[{Boubaker et~al.}(2013){Boubaker, Elouedi, \andnamec{}
  Lefevre}]{Boubaker13}
Boubaker, J., Z.~Elouedi, \andname{} E.~Lefevre (2013).
\newblock Conflict management with dependent information sources in the belief
  function framework.
\newblock In {\em The 14th IEEE International Symposium on Computational
  Intelligence and Informatics, CINTI 2013}, Budapest, Hongrie.

\bibitem[{Chebbah et~al.}(2012a){Chebbah, Martin, \andnamec{} Ben{
  }Yaghlane}]{Chebbah12a}
Chebbah, M., A.~Martin, \andname{} B.~Ben{ }Yaghlane (2012a).
\newblock About sources dependence in the theory of belief functions.
\newblock In {\em Belief Functions}, pp.\  239--246.

\bibitem[{Chebbah et~al.}(2012b){Chebbah, Martin, \andnamec{} Ben{
  }Yaghlane}]{Chebbah12b}
Chebbah, M., A.~Martin, \andname{} B.~Ben{ }Yaghlane (2012b).
\newblock Positive and negative dependence for evidential database enrichment.
\newblock In {\em IPMU (3)}, pp.\  575--584.

\bibitem[{Dempster}(1967){Dempster}]{Dempster67a}
Dempster, A.~P. (1967).
\newblock Upper and {L}ower probabilities induced by a multivalued mapping.
\newblock {\em Annals of Mathematical Statistics\/}~{\em 38}, 325--339.

\bibitem[{Den{\oe}ux}(2006){Den{\oe}ux}]{Denoeux06a}
Den{\oe}ux, T. (2006).
\newblock The cautious rule of combination for belief functions and some
  extensions.
\newblock In {\em International Conference on Information Fusion}, Florence,
  Italy.

\bibitem[{Den{\oe}ux}(2008){Den{\oe}ux}]{Denoeux08a}
Den{\oe}ux, T. (2008).
\newblock Conjunctive and disjunctive combination of belief functions induced
  by nondistinct bodies of evidence.
\newblock {\em Artificial Intelligence\/}~{\em 172}, 234--264.

\bibitem[{Dubois \andnamec{} Prade}(1988){Dubois \andnamec{} Prade}]{Dubois88}
Dubois, D. \andname{} H.~Prade (1988).
\newblock Representation and combination of uncertainty with belief functions
  and possibility measures.
\newblock {\em Computational Intelligence\/}~{\em 4}, 244--264.

\bibitem[{Elouedi \andnamec{} Mellouli}(1998){Elouedi \andnamec{}
  Mellouli}]{Elouedi98}
Elouedi, Z. \andname{} K.~Mellouli (1998).
\newblock Pooling dependent expert opinions using the theory of evidence.
\newblock In {\em Seventh Information Processing and Management of Uncertainty
  in Knowledge-Based System (IPMU'98)}, Volume~I, pp.\  32--39.

\bibitem[{Hsia}(1991){Hsia}]{Hsia91}
Hsia, Y.-T. (1991).
\newblock Characterizing belief with minimum commitment.
\newblock In {\em IJCAI}, pp.\  1184--1189.

\bibitem[{Jousselme et~al.}(2001){Jousselme, Grenier, \andnamec{}
  Boss\'e}]{Jousselme01a}
Jousselme, A.-L., D.~Grenier, \andname{} E.~Boss\'e (2001).
\newblock A new distance between two bodies of evidence.
\newblock {\em Information Fusion\/}~{\em 2}, 91--101.

\bibitem[{Martin}(2010){Martin}]{Martin10a}
Martin, A. (2010).
\newblock Le conflit dans la th{\'e}orie des fonctions de croyance.
\newblock In {\em Actes Extraction et gestion des connaissances (EGC'2010)},
  Hammamet, Tunisia, pp.\  655--666.

\bibitem[{Martin \andnamec{} Osswald}(2007){Martin \andnamec{}
  Osswald}]{Martin07a}
Martin, A. \andname{} C.~Osswald (2007).
\newblock Une nouvelle r\`egle de combinaison r\'epartissant le conflit -
  applications en imagerie sonar et classification de cibles radar.
\newblock {\em Traitement du Signal\/}~{\em 24\/}(2), 71--82.

\bibitem[{Mercier}(2006){Mercier}]{Mercier06a}
Mercier, D. (2006).
\newblock {\em Fusion d'informations pour la reconnaissance automatique
  d'adresses postales dans le cadre de la théorie des fonctions de croyance}.
\newblock Ph.\ D. thesis, Université de Technologie de Compiègne.

\bibitem[{Murphy}(2000){Murphy}]{Murphy00a}
Murphy, C. (2000).
\newblock Combining belief functions when evidence conflicts.
\newblock {\em Decision Support Systems\/}~{\em 29}, 1--9.

\bibitem[{Shafer}(1976){Shafer}]{Shafer76}
Shafer, G. (1976).
\newblock {\em A mathematical theory of evidence}.
\newblock Princeton University Press.

\bibitem[{Smets}(1990){Smets}]{Smets90a}
Smets, P. (1990).
\newblock The {C}ombination of {E}vidence in the {T}ransferable {B}elief
  {M}odel.
\newblock {\em IEEE Transactions on Pattern Analysis and Machine
  Intelligence\/}~{\em 12\/}(5), 447--458.

\bibitem[{Smets}(1993){Smets}]{Smets93a}
Smets, P. (1993).
\newblock {B}elief {F}unctions: the {D}isjunctive {R}ule of {C}ombination and
  the {G}eneralized {B}ayesian {T}heorem.
\newblock {\em International Journal of Approximate Reasoning\/}~{\em 9},
  1--35.

\bibitem[{Smets}(2005){Smets}]{Smets05b}
Smets, P. (2005).
\newblock Decision making in the {TBM}: the necessity of the pignistic
  transformation.
\newblock {\em International Journal of Approximate Reasonning\/}~{\em 38},
  133--147.

\bibitem[{Smets \andnamec{} Kennes}(1994){Smets \andnamec{} Kennes}]{Smets94a}
Smets, P. \andname{} R.~Kennes (1994).
\newblock The {T}ransferable {B}elief {M}odel.
\newblock {\em Artificial Intelligent\/}~{\em 66}, 191--234.

\bibitem[{Smets \andnamec{} Kruse}(1997){Smets \andnamec{} Kruse}]{Smets97c}
Smets, P. \andname{} R.~Kruse (1997).
\newblock {\em The transferable belief model for belief representation}, pp.\
  343--368.
\newblock Boston: Kluwer Academic Publishers.

\bibitem[{Yager}(1987){Yager}]{Yager87}
Yager, R.~R. (1987).
\newblock On the {D}empster-{S}hafer {F}ramework and {N}ew {C}ombination
  {R}ules.
\newblock {\em Information Sciences\/}~{\em 41}, 93--137.

\end{thebibliography}

\end{document}